\documentclass[11pt]{article} %
\usepackage{rldmsubmit,palatino}
\usepackage{graphicx}

\usepackage{amsmath}                    %
\usepackage{amssymb}                    %
\usepackage{amsbsy}                     %
\usepackage{amsfonts}                   %
\usepackage{amsthm}                     %
\usepackage{bbold}                      %
\usepackage{bm}                         %
\usepackage{physics}                    %
\usepackage{placeins}                   %
\usepackage{textgreek}                  %
\usepackage{xspace}                     %
\usepackage{graphicx}
\graphicspath{ {figure/} }
\usepackage{booktabs}

\usepackage[tworuled,nofillcomment]{algorithm2e}

\usepackage{cleveref}

\title{Incrementally Learning Functions of the Return}

\author{
Brendan Bennett\thanks{ Corresponding author.} \\
Department of Computer Science\\
University of Alberta\\
Edmonton, AB T6G 2E8 \\
\texttt{babennet@ualberta.ca} \\
\And
Wesley Chung \\
Department of Computer Science\\
University of Alberta\\
Edmonton, AB T6G 2E8 \\
\texttt{wchung@ualberta.ca} \\
\AND
Muhammad Zaheer \\
Department of Computer Science\\
University of Alberta\\
Edmonton, AB T6G 2E8\\
\texttt{mzaheer@ualberta.ca} \\
\And
Vincent Liu \\
Department of Computer Science\\
University of Alberta\\
Edmonton, AB T6G 2E8\\
\texttt{vliu1@ualberta.ca} \\
}

\usepackage[backend=biber,style=ieee,backref=true,sortcites=true,sorting=nyt]{biblatex}

\theoremstyle{definition}

\crefname{algocf}{alg.}{algs.}
\Crefname{algocf}{Algorithm}{Algorithms}

\usepackage{upgreek}
\DeclareRobustCommand{\mathup}[1]{\begingroup\changegreek\mathrm{#1}\endgroup}

\makeatletter
\def\changegreek{\@for\next:={%
  alpha,beta,gamma,delta,epsilon,zeta,eta,theta,kappa,lambda,mu,nu,xi,pi,rho,sigma,%
  tau,upsilon,phi,chi,psi,omega,varepsilon,vartheta,varpi,varrho,varsigma,varphi}%
  \do{\expandafter\let\csname\next\expandafter\endcsname\csname up\next\endcsname}}
\def\changegreekbf{\@for\next:={%
  alpha,beta,gamma,delta,epsilon,zeta,eta,theta,kappa,lambda,mu,nu,xi,pi,rho,sigma,%
  tau,upsilon,phi,chi,psi,omega,varepsilon,vartheta,varpi,varrho,varsigma,varphi}%
  \do{\expandafter\def\csname\next\expandafter\endcsname\expandafter{%
    \expandafter\bm\expandafter{\csname up\next\endcsname}}}}
\makeatother

\newenvironment{eqalign}
    {\begin{equation}\begin{aligned}}
    {\end{aligned}\end{equation}}

\newenvironment{eqalign*}
    {\begin{equation*}\begin{aligned}}
    {\end{aligned}\end{equation*}}

\newcommand{\myhat}[1]
    {\ensuremath{\bm\hat{#1}}}

\newcommand{\reals}{\ensuremath{\mathbb{R}}}                    %
\renewcommand{\vec}[1]{{\boldsymbol{\mathbf{\mathup{#1}}}}}     %
\newcommand{\zeros}{\vec{0}}                                    %

\let\epsilon\varepsilon

\newcommand{\eqdef}{\ensuremath{\stackrel{\text{\tiny{def}}}{=}}}

\newcommand{\Exp}{\ensuremath{\mathop{\mathbb{E}}}}     %
\newcommand{\Var}{\ensuremath{\mathop{\text{Var}}}}     %

\newcommand{\states}{\ensuremath{\mathcal{S}}}

\begin{document}

\maketitle

\begin{abstract}
	Temporal difference methods enable efficient estimation of value functions in reinforcement learning in an incremental fashion, and are of broader interest because they correspond learning as observed in biological systems.
	Standard value functions correspond to the expected value of a sum of discounted returns.
	While this formulation is often sufficient for many purposes, it would often be useful to be able to represent \emph{functions} of the return as well.
	Unfortunately, most such functions cannot be estimated directly using TD methods.
	We propose a means of estimating functions of the return using its moments, which can be learned online using a modified TD algorithm.
	The moments of the return are then used as part of a Taylor expansion to approximate analytic functions of the return.

\end{abstract}


\acknowledgements{
	We are deeply grateful to the Richard S. Sutton and Martha White foundations for their support of wayward graduate students.
}

\startmain %

\section{Introduction}
\label{sec:introduction}

In the context of Markov Decision Processes, the value for state $s$ under policy $\pi$ is given by:
\begin{eqalign}
	\label{eq:defn-value-function}
	v_{\pi}(s)
	&\eqdef \Exp \qty[ G_{t} | S_{t} = s, A_{t+k} \sim \pi(S_{t+k}) ], \quad k = 0, 1, 2, \ldots
\end{eqalign}
where we define the return $G_{t} \eqdef \sum_{n=0}^{\infty} \gamma^{n} R_{t+n+1}$, the sum of discounted future rewards.
Many problems can be modelled using MDPs, and so by learning the value function we become able to predict the results of a given policy.

While this formulation captures many quantities of interest, sometimes what we wish to predict cannot be expressed as discounted sums of rewards.
For example, financiers evaluate potential investments in terms of both the expected profit as well as the risk they would be undertaking.

Whereas the standard value function only reflects the expected return, we can instead take inspiration from economics and consider a \emph{utility} function such as
\begin{eqalign*}
	u(s)
	= \mathbb{E}[G_{t} | S_{t} = s] + \beta \Var [G_{t} | S_{t} = s]
	= v(s) + \beta \Var [G_{t} | S_{t} = s]
\end{eqalign*}
A negative value of $\beta$ models risk-aversion, while a positive value corresponds to risk-seeking; when $\beta$ is zero, the utility is identical to the expected value.

Unfortunately, the variance---and most other functions of the return---are \emph{not} directly amenable to standard RL methods, as it cannot be written in the form of a standard return, \textit{i.e.} as a simple recursive equation.
This is generally the case for any function of the return that is not linear.

In this paper, we investigate a method for learning the moments of the return in a way that is compatible with temporal difference (TD) learning.
This in turn allows us to estimate \emph{nonlinear} functions of the return via Taylor approximation.
The algorithm we describe is patterned off of standard TD learning, although our methods are easily adaptable to other algorithms if so desired. 
Further examples and some experiments illustrating the usefulness and validity of our approach are provided.

\section{Moments of the Return}

As a first example of more general approximation targets, we consider the problem of estimating the moments of the return\footnote{
	Recall that for a random variable $X$, the $n$-th moment of $X$ (if it exists) is defined to be $M_{n}(X) \eqdef \Exp[X^{n}]$, and the $n$-th central moment is $C_{n}(X) \eqdef \Exp[(X - \Exp[X])^{n}]$.
	The first moment of $X$ is just its expected value, and the second central moment is more commonly termed the variance.
}.
In general, the formula for the $n$-th power of the return is:
\begin{eqalign}
	\label{eq:defn-return-nth-power}
	G_{t}^{n}
	&\eqdef ( R_{t+1} + \gamma G_{t+1} )^{n}
	= \sum_{k=0}^{n-1} \binom{n}{k} \gamma^{k} R_{t+1}^{n-k} G_{t+1}^{k} +  \gamma^{n} G_{t+1}^{n}
	\\
	&= R_{t+1}^{(n)} + \gamma^{n} G_{t+1}^{n}
\end{eqalign}
Where we use $R_{t+1}^{(n)} \eqdef \sum_{k=0}^{n-1} \binom{n}{k} \gamma^{k} R_{t+1}^{n-k} G_{t+1}^{k}$.
So the $n$-th moment is just $\Exp [ R_{t+1}^{(n)} ] + \gamma^{n} \Exp[G_{t+1}^{n}]$.
Superficially, this has the appearance of a Bellman equation, but note that due to the terms involving a product of $R_{t+1}$ and $G_{t+1}$ we can no longer leverage the Markov property to update from a single transition.

However (as noted by \cite{Sobel1982,Tamar2016,White2016c} among others) it is possible to make an approximation to get a more appropriate target.
If we have an estimate of the value function for the first $n-1$ moments, written $v_{n}(\cdot)$, with $v_{n}(s) \approx \Exp_{\pi,s} [G_{t}^{n}]$, then we can define a new approximation target for the next moment, $\myhat{G}^{n}$, via:
\begin{eqalign}
	\label{eq:defn-hat-return-nth-power}
	\myhat{G}^{n}
	&\eqdef \qty( \sum_{k=0}^{n-1} \binom{n}{k} \gamma^{k} R_{t+1}^{n-k} v_{k}(S_{t+1}) ) + \gamma^{n} \myhat{G}_{t+1}^{n}
	= \myhat{R}_{t+1}^{(n)} + \gamma^{n} \myhat{G}_{t+1}^{n}
\end{eqalign}
Where we simplify the notation by defining $v_{0}:\states \rightarrow \reals$ such that $v_{0}(s) = 0$ if $s$ is a terminal state and $1$ otherwise.
Here, $\myhat{R}_{t+1}^{(n)}$ is given by:
\begin{eqalign}
	\label{eq:defn-hat-reward}
	\myhat{R}_{t+1}^{(n)}
	&\eqdef \sum_{k=0}^{n-1} \binom{n}{k} \gamma^{k} R_{t+1}^{n-k} v_{k}(S_{t+1})
\end{eqalign}
This new approximation target\footnote{
	Using a proxy in place of a quantity that is more difficult to approximate is a common tactic in machine learning.
	In fact, the $\lambda$-return (the approximation target of TD($\lambda$) and related methods) is similar in some respects.
	Given by $G_{t}^{\lambda} \eqdef R_{t+1} + \gamma ( \lambda G_{t+1}^{\lambda} + (1 -\lambda) v(S_{t+1}) )$, it \emph{also} requires a value function in order to be properly specified.
}
is similar to \cref{eq:defn-return-nth-power} except that we use $v_{n}(S_{t+1})$ in place of $G_{t+1}^{n}$ in the binomial sum.
It has the form of a Bellman equation and can be learned via TD methods (see \cite{Munos1999,Sobel1982,Tamar2016,White2016c,bennett2019thesis} for further elaboration).

Ideally, we would have $\myhat{G}^{n} = G^{n}$---this is the case if the estimated value functions $v_{1}, v_{2}, \ldots, v_{n-1}$ are equivalent to their corresponding ``true" value functions.
When the value functions are inexact (as under function approximation), $\myhat{G}^{n}$ is still well-defined, but is likely to be a biased estimate of the true $G^{n}$.

If desired, an approximation for the \emph{central} moments $c_{n}(s)$ can then be calculated from the estimates via:
\begin{eqalign}
	\label{eq:central-moments-conversion}
	c_{n}(s)
	&= \Exp_{\pi,s}[ (G_{t} - \Exp_{\pi,s} [ G_{t} ])^{n} ]
	= \sum_{k=0}^{n} \binom{n}{k} (-1)^{n-k} v_{k}(s) v_{1}^{n-k}(s)
\end{eqalign}
In \Cref{alg:td-moment-estimation} we provide an example implementation for estimating the higher moments of the return\footnote{
	A more detailed treatment that includes bootstrapping and off-policy extensions for the second moment is described in~\cite{White2016c,Sherstan2018}.
} based on the Semi-Gradient TD($\lambda$) algorithm from~\cite{sutton2018reinforcement} (pg.293).

{
\setlength{\algomargin}{2em}
\begin{algorithm}[h]
	\DontPrintSemicolon
	\SetKwComment{comment}{$\triangleright$\ }{}
	\SetCommentSty{emph}
	\SetKw{Input}{input :}
	\SetKw{Output}{output :}
	\SetKw{Parameters}{parameters : }
	\SetAlgoLined

	\Input{the policy $\pi$ to be evaluated} \;
	\Input{A set of $n$ differentiable functions $\{ v_{i} \}_{i=1}^{n}$ parameterized by their respective weights $\{ \vec{w} \}_{i=1}^{n}$ with $\vec{w}_{i} \in \reals^{d}$, such that $v_{i} : \states \rightarrow \reals$ and $v_{i}(\text{terminal}) = 0$.
	} \;
	\Parameters{
	A set of step-sizes $\{ \alpha_{i} \}_{i=1}^{n}$ with $\alpha_{i} \in (0, 1)$;
	a set of trace decay rates $\{ \lambda \}_{i=1}^{n}$ with $\lambda_{i} \in [0, 1]$.
	} \;
	\BlankLine

	\lForEach{$k = 1, \ldots, n$}{Initialize $\vec{w}_{k}$ arbitrarily}
	\For{each episode}{
		Initialize $s$      \comment*{Initial state}
		\lForEach(\comment*[f]{set traces to zero}){$k = 1, \ldots, n$}{$\vec{z}_{k} \gets \zeros$}
		\While{$s$ is not terminal}{
			Choose $a \sim \pi(\cdot | s)$\;
			Take action $a$, observe $r, s'$ \;
			\For{$k = n, \ldots, 1$}{
				$\vec{z}_{k} \gets \gamma^{k} \lambda_{k} + \nabla v_{k}(s)$ 
				\comment*{update traces}
				$r_{k} \gets \sum_{\ell=0}^{k-1} \binom{k}{\ell} \gamma^{\ell} r^{k - \ell} v_{\ell}(s')$
				\comment*{``reward" from \cref{eq:defn-hat-reward}}
				$\delta \gets r_{k} + \gamma^{k} v_{k}(s') - v_{k}(s)$ \;
				$\vec{w}_{k} \gets \vec{w}_{k} + \alpha_{k} \delta \vec{z}_{k}$ \;
			}
			$s \gets s'$ \;
		}
	}
	\caption{Estimating $n$ moments of the return for a policy using TD($\lambda$)}
	\label{alg:td-moment-estimation}
\end{algorithm}
}

It is possible to prove convergence of \Cref{alg:td-moment-estimation} with linear function approximation and $\lambda=0$.
Due to the space constraints of this paper, we do not provide those, but refer the interested reader to~\cite{Tamar2016,bennett2019thesis}. %

\section{Approximating Functions of the Return via Taylor Expansions}

We now suggest some methods for estimating functions of the return.
Consider an analytic function
$f: \reals \rightarrow \reals$ and define $u_{\pi} : \states \rightarrow \reals$ by
\begin{eqalign}
	u_{\pi}(s)
	\eqdef \Exp_{\pi} [ f(G_{t}) | S_{t} = s ]
	= \Exp_{\pi, s} [ \widetilde{G}_{t} ]
\end{eqalign}
where $\widetilde{G}_t = f(G_t)$.
We wish to approximate this function with some $u(\cdot)$ so that $u_{\pi} \approx u$.

Recall that an analytic function can be expressed by the Taylor expansion about some point $a$ by:
\begin{eqalign}
	f(x)
	&= \sum_{n=0}^{\infty} \frac{ f^{(n)}(a) }{ n! } (x - a)^{n}
	= f(a) + f'(a) (x - a) + \frac{1}{2!} f''(a) (x - a)^{2} + \cdots
\end{eqalign}
with $f^{(n)}$ denoting the $n$-th derivative of $f$.
So long as $x$ lies within the radius of convergence for the Taylor expansion at $a$, the expansion is exact and (tautologically) convergent\footnote{
	The radius of convergence depends on both $f$ and $a$, but one could either rescale the function or the rewards to ensure the expansion's validity.
}.

This can be used to express the expected value of a function of a random variable, in this case, the return:
\begin{eqalign}
	\Exp_{\pi}[ f( G_{t} ) | S_{t} = s ]
	&= \Exp_{\pi,s} \qty[ \sum_{n=0}^{\infty} \frac{ f^{(n)}(a) }{ n! } (G_{t} - a)^{n} ]
	= \sum_{n=0}^{\infty} \frac{ f^{(n)}(a) }{ n! } \Exp_{\pi,s} \qty[ (G_{t} - a)^{n} ]
\end{eqalign}
where we use the linearity of expectation and the fact that $f^{(n)}(a)$ is a effectively a constant (given $a$) to rewrite the sum.
There are two ``obvious" choices for $a$: either the origin or the mean, $v_{1}(s)$.
In the case of $a = 0$ this just becomes
\begin{eqalign}
	u(s)
	&= \sum_{n=0}^{\infty} \frac{ f^{(n)}(0) }{ n! } \Exp_{\pi,s} \qty[ G_{t}^{n} ]
	\approx \sum_{n=0}^{\infty} \frac{ f^{(n)}(0) }{ n! }  v_{n}(s)
\end{eqalign}
Whereas for $a = v_{1}(s)$ we get
\begin{eqalign}
	u(s)
	&= \sum_{n=0}^{\infty} \frac{ f^{(n)}(v_{1}(s)) }{ n! } \Exp_{\pi,s} \qty[ (G_{t} - v_{1}(s))^{n} ]
	\approx \sum_{n=0}^{\infty} \frac{ f^{(n)}( v_{1}(s) ) }{ n! }  c_{n}(s)
\end{eqalign}
where $c_{n}(\cdot)$ can be computed from \cref{eq:central-moments-conversion} as needed.

We remark that this technique is surprisingly effective, but can break down if the higher moments of $G_{t}$ dominate the sum because the estimates for those moments tend to be less accurate.

\section{Experiments}
\label{sec:experiments}

\begin{figure}[h!]
	\centering
	\includegraphics[width=0.9\linewidth]{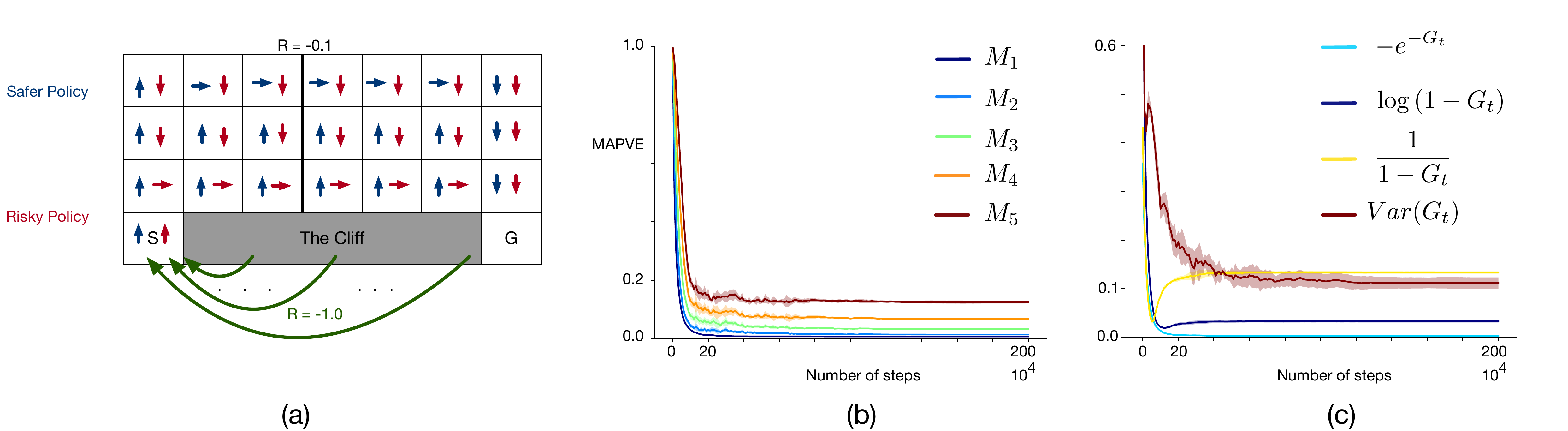}
	\caption{(a) The cliff walk domain. Risky policy is indicated by red arrows, and safe policy is indicated by blue arrows. (b) The prediction error for the first five moments of the return. (c) The prediction error for different functions of return.}
	\label{fig:cliff}
\end{figure}

We use a variant of the Cliff Walk domain \cite{sutton2018reinforcement} as depicted in Figure \ref{fig:cliff} (a). 
The agent can move up, down, left, or right, but with a small probability of ending up in a square adjacent to the one it intended to travel to.
The agent receives a small negative reward for each timestep until it reaches the goal.
If the agent enters one of the ``cliff" cells, it receives a large negative reward and is transferred back to the start position.

We then define two policies: a risky one which takes the short path near the cliff and a safe one which takes a longer path avoiding the cliff.
Extensive Monte-Carlo rollouts were used to compute the true value functions for the modified returns and we utilize the mean absolute percentage value error, $MAPVE = \frac{1}{|S|} \sum_{s \in S} \frac{|\hat{v}(s) - v(s)|}{|v(s)|}$ where $\hat{v}(s)$ is the approximated value and $v(s)$ is the true value function for the modified returns.

To demonstrate the validity of our approach, we estimate the first five moments of the return using \cref{alg:td-moment-estimation} on the Safe policy with $\lambda=0$. 
Figure \ref{fig:cliff} (b) shows that we are indeed able to approximate the moments accurately, with their errors converging towards zero.
We then use these estimated moments to form a Taylor approximation to a few functions of the return.
We find that the Taylor approximation is also reasonably accurate.

Next, we show that, by considering functions of the return, we may prefer different policies when compared to just using the value function.
Concretely, we use $f(G_t) = -e^{-G_t}$ which magnifies negative returns. 
In the Cliff Walk domain, this tends to penalize risky policies that have a greater chance of falling off the cliff, more than offsetting the benefit of a faster route. 
We compare the two policies (see \cref{fig:cliff}.a) mentioned earlier.
Table~\ref{tab:policy-utility} shows the estimates learned by our algorithm, for both the expected return and its transformation $\mathbb{E}_{\pi}[e^{-G_{t}}]$.

\begin{table}[ht]
	\centering
	\begin{tabular}{c|cc}
		{}           & $\mathbb{E}_{\pi}[G_{0}]$ & $\mathbb{E}_{\pi}[e^{-G_{0}}]$ \\ \hline
		Safe Policy  & -1.45           & \textbf{-4.60}     \\
		Risky Policy & \textbf{-1.30}  & -7.25              \\
	\end{tabular}
	\caption{
	   Estimates learned for the start state (t=0) for the different policies.
	}
	\label{tab:policy-utility}
\end{table}
From Table~\ref{tab:policy-utility}, we observe that different policies are preferable depending on how we define the utility function.
If utility is equivalent to value, the Risky policy has the advantage, since it tends to reach the goal faster in spite of occasionally plummeting off the cliff.
But if we instead evaluate the policies in terms of the transformed return, the Safe policy is better.
We conclude that transforming the return can be an effective strategy for modifying the preferences of agents.

\section{Summary}

In this brief report, we have indicated some situations where the value function alone might not capture salient aspects of a task. 
To some extent, this can be ameliorated by instead using a function of the return.
We proposed a method that allows for estimating functions of the return while retaining the advantages of temporal difference methods: fast, incremental learning with low resource requirements.

From a theoretical perspective, predictions (in the form of value functions) are fundamental to reinforcement learning, so a means of making more expressive predictions can only be advantageous.
The most straightforward use case would be improving an agent's performance by more accurately specifying what constitutes desirable behavior, such as in the Cliff Walk experiment described in \cref{sec:experiments}.

There are also some less obvious applications.
Learning value functions can assist agents in many ways, being used to explicitly refine the agent's knowledge about its environment~\cite{dayan1993improving}, or as an auxiliary task that improves the representation when deep learning techniques are employed~\cite{jaderberg2016reinforcement}.

On a more speculative note, our ideas might also have relevance for theories of biological cognition.
We note that there is evidence for TD-like mechanisms in the human brain~\cite{o2003temporal}, and that it matches well with models of animal learning.
Perhaps there are some neural correlates of moment estimation, or even something corresponding to estimating functions of the return.

While we limited our investigation to TD methods, the approach we suggest is very general, and could be applied with other algorithms instead of \Cref{alg:td-moment-estimation}.
By extending what we can express in terms of value functions, this allows for greater flexibility in making predictions and setting objectives, without requiring new algorithms or modifications to the ``natural" reward function of a task.

{\small
\printbibliography
}
\end{document}